\documentclass{article}
\usepackage{ijcai22}

\usepackage{times}
\usepackage{soul}
\usepackage{url}
\usepackage[colorlinks,linkcolor=red]{hyperref}
\usepackage[utf8]{inputenc}
\usepackage[small]{caption}
\usepackage{graphicx}
\usepackage{amsmath}
\usepackage{amsthm}
\usepackage{booktabs}
\usepackage{algorithm}
\usepackage{algorithmic}
\usepackage{hyperref}
\title{DecisionHoldem: Safe Depth-Limited Solving With Diverse Opponents for Imperfect-Information Games}

\author{
	Qibin Zhou$^{1}$\and
	Dongdong Bai$^1$\footnote{Dongdong Bai and Qibin Zhou contribute equally. }\and
	Junge Zhang$^1$\footnote{Corresponding Author}\and
	Fuqing Duan$^2$\and
	Kaiqi Huang$^1$\\
	\affiliations
	$^1$Institute of Automation, Chinese Academy of Sciences, Beijing, China\\
	$^2$Beijing Normal University, Beijing, China\\
	\emails
	zqbagent@gmail.com,
	baidongdong@nudt.edu.cn,
	jgzhang@nlpr.ia.ac.cn,
	fqduan@bnu.edu.can,
	kqhuang@nlpr.ia.ac.cn
}

\begin{document}

\maketitle

\begin{abstract}
	
An imperfect-information game is a type of game with asymmetric information. It is more common in life than perfect-information game. Artificial intelligence (AI) in imperfect-information games, such like poker, has made considerable progress and success in recent years. The great success of superhuman poker AI, such as Libratus and Deepstack, attracts researchers to pay attention to poker research. However, the lack of open-source code limits the development of Texas hold'em AI to some extent. This article introduces DecisionHoldem, a high-level AI for heads-up no-limit Texas hold'em with safe depth-limited subgame solving by considering possible ranges of opponent's private hands to reduce the exploitability of the strategy. Experimental results show that DecisionHoldem defeats the strongest openly available agent in heads-up no-limit Texas hold'em poker, namely Slumbot, and a high-level reproduction of Deepstack, viz, Openstack, by more than 730 mbb/h (one-thousandth big blind per round) and 700 mbb/h. Moreover, we release the source codes and tools of DecisionHoldem to promote AI development in imperfect-information games.

\end{abstract}

\section{Introduction}

The success of AlphaGo~\cite{silver2016mastering} has led to increasing attention to the study of game decision-making~\cite{brown2018superhuman,moravvcik2017deepstack,brown2018depth,brown2019superhuman}. Unlike perfect-information games, such as Go,  real-world problems are mainly imperfect-information games. The hidden knowledge in poker games (i.e., private cards) corresponds to the real world's imperfect-information. Research on poker artificial intelligence (AI) can provide means to deal with problems in life, such as financial market tracking and stock forecasting.

Research in imperfect-information games, particularly poker AI~\cite{brown2018superhuman,moravvcik2017deepstack,brown2018depth,brown2019superhuman}, has made considerable progress in recent years. Texas hold'em is one of the most popular poker game in the world. It is an excellent benchmark for studying the game theory and technology in imperfect-information games because of the following three factors. First, Texas hold'em is a typical imperfect-information game. Before the game, two private hands invisible to the opponent are distributed to each player. Players should predict the opponents' private hands during decision-making based on the opponents' historical actions, which makes Texas hold'em obtain the characteristics of deception and anti-deception. Second, the complexity of the Texas hold'em game is enormous. The decision-making space for heads-up no-limit Texas hold'em (HUNL) exceeds $10^{160}$~\cite{johanson2013measuring}. In addition, Texas hold'em has simple rules and moderate difficulty, which considerably facilitates the verification of algorithms by researchers.

After decades of research, the poker AI DeepStack developed by Matej Moravčík et al.~\cite{moravvcik2017deepstack} and Libratus developed by Noam Brown and Tuomas Sandholm~\cite{brown2018superhuman} successively defeat human professional players in 2017. This event affirms the breakthrough for HUNL. Subsequently, the poker AI Pluribus, also constructed by Noam Brown and Tuomas Sandholm~\cite{brown2019superhuman}, defeats the human professional players in six-man no-limit Texas hold'em. Although \emph{Science} magazine has published the poker AI mentioned above~\cite{brown2018superhuman,moravvcik2017deepstack}, the relevant code and main technical details have not been made public.

In addtion, considerable poker AI progress~\cite{Brown2017DynamicTA} 
\cite{Hartley2017MultiAgentCR}
\cite{Brown2019SolvingIG}
\cite{Schmid2019VarianceRI}
\cite{Farina2019RegretCC}
\cite{Farina2019StablePredictiveOC}
\cite{Farina2019OptimisticRM}
\cite{Li2020DoubleNC} 
is only tested in games with small decision space, such as Leduc hold'em and Kuhn Poker. These algorithms may not work well when applied to large-scale games, such as Texas hold'em.

\begin{table*}[ht]
	\centering
	\begin{tabular}{ccccc}
		\toprule
		\textbf{Round} & \textbf{Number of Abstract Hands} & \textbf{1st $\sim$ 2nd Actions} & \textbf{3rd $\sim$ 5th Actions} &  \textbf{Remaining Actions} \\
		\midrule
		Pre-Flop & 169 & F,\enspace C,\enspace 0.5P,\enspace P,\enspace 2P,\enspace 4P,\enspace A & F,\enspace C,\enspace P,\enspace 2P,\enspace 4P,\enspace A & F,\enspace C,\enspace  A \\
		Flop & 50,000 & F,\enspace C,\enspace 0.5P,\enspace P,\enspace 2P,\enspace 4P,\enspace A & F,\enspace C,\enspace P,\enspace 2P,\enspace 4P,\enspace A & F,\enspace C,\enspace  A \\
		Turn & 5,000 & F,\enspace C,\enspace 0.5P,\enspace P,\enspace 2P,\enspace 4P,\enspace A & F,\enspace C,\enspace P,\enspace 2P,\enspace 4P,\enspace A & F,\enspace C,\enspace  A \\
		River & 1,000 & F,\enspace C,\enspace 0.5P,\enspace P,\enspace 2P,\enspace 4P,\enspace A & F,\enspace C,\enspace P,\enspace 2P,\enspace 4P,\enspace A & F,\enspace C,\enspace  A \\
		\bottomrule
	\end{tabular}
	\caption{The number of abstract hands and actions available for each round (pre-flop, flop, turn and, river) of DecisionHoldem on HUNL. F, C, 0.5P, P, 2P, 4P, and A represent Fold, Call, 0.5 Pot size, 1.0 Pot size, 2.0 Pot size, 4.0 pot size, and all-in, respectively.}
	\label{table1}
\end{table*}

\begin{figure}[ht]
	\centering
	\includegraphics[width=1\linewidth]{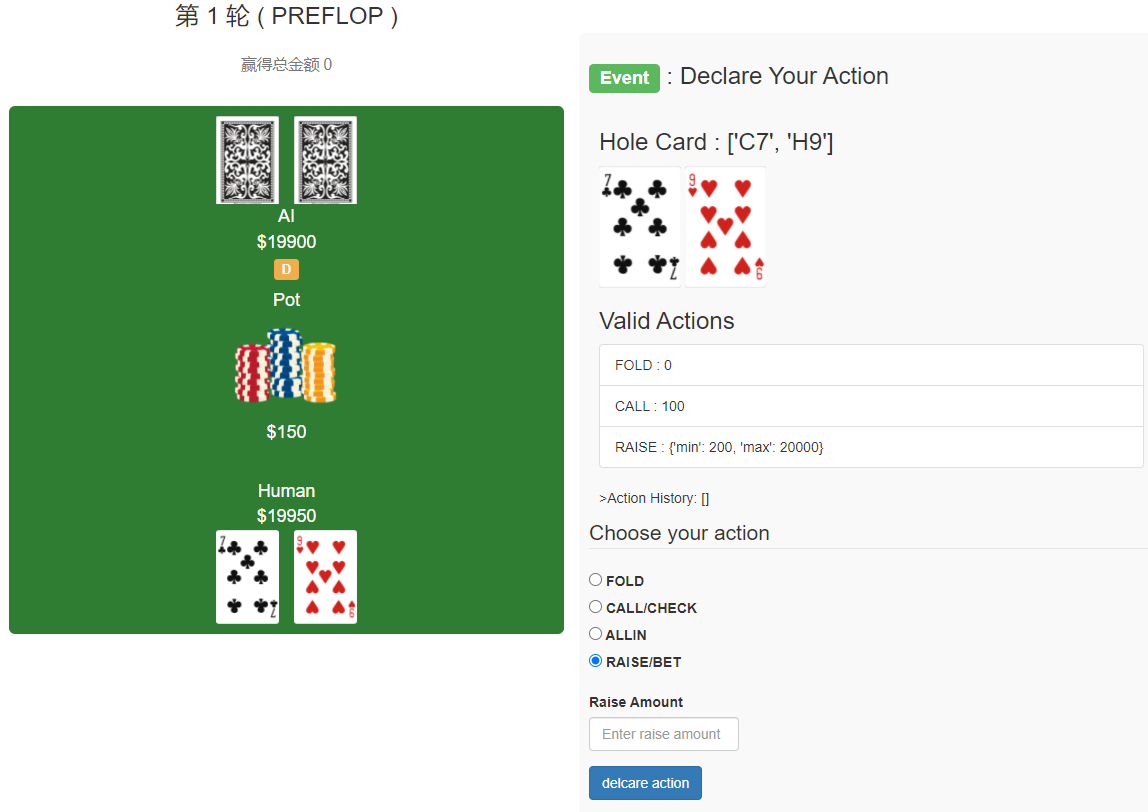}\caption{Demonstration of AI and human confrontation.}
	\label{figure1}
\end{figure}

In this paper, we propose a safe depth-limited subgame solving algorithm with diverse opponents. To evaluate the algorithm's performance, we achieve a high-performance and high-efficiency poker AI based on it, namely DecisionHoldem. Experiments show that DecisionHoldem defeats the strongest public poker AI, such as Slumbot\footnote{www.slumbot.com} (champion of 2018 Annual Computer Poker Competition [ACPC]) and OpenStack (a reproduction of DeepStack built-in OpenHoldem~\cite{li2020openholdem}\footnote{holdem.ia.ac.cn}, by a big margin. Meanwhile, we release DecisionHoldem's source code, and tools for playing against the Slumbot and OpenHoldem~\cite{li2020openholdem}. In addition, we also provide a platform to play DecisionHoldem with humans (as in Figure \ref{figure1}\footnote{https://github.com/ishikota/PyPokerGUI}). Our code is available at \url{https://github.com/AI-Decision/DecisionHoldem}.

\section{Methods}
In this study, we use the counterfactual regret minimization (CFR) algorithm~\cite{zinkevich2007regret}, the primary way of the Texas hold'em AI, and combine it with safe depth-limited subgame solving to achieve the high-performance and high-efficiency poker AI --- DecisionHoldem. DecisionHoldem is mainly composed of two parts, namely the blueprint strategy and the real-time search part. 

In the blueprint strategy part, we partially follow the idea of Libratus but adjusted the parameters of the abstract number of actions and hands. The abstract parameters of DecisionHoldem's hands and actions are shown in Table \ref{table1}. DecisionHoldem first employs the hand abstraction technique and action abstraction to obtain an abstracted game tree. Then, we use the linear CFR algorithm \cite{brown2019solving} iteration on the abstracted game tree to calculate blueprint strategy on a workstation with 48 core CPUs for about 3 $\sim$ 4 days with approximately 200 million iterations. The total computing power cost is about 4,000 core hours.

In the real-time search part, we propose a safer depth-limited subgame solving algorithm than modicum's~\cite{brown2018depth} on subgame solving by considering diverse opponents for off-tree nodes. Since the opponent's private hand range reflects the opponent's play style and strategy, we propose a safe depth-limited subgame solving method by explicitly modeling diverse opponents with different ranges. This algorithm can refine the degree of subgame strategy without worsening the exploitability compared with the blueprint strategy. That is to say, safe depth-limited solving with diverse opponents can significantly enhance the AI decision-making level and ability with changeable challenges. Our subsequent articles will introduce the details of the algorithm.

\section{Experiments and Results}

DecisionHoldem plays against Slumbot and OpenStack~\cite{li2020openholdem} to test its capability. Slumbot is the champion of the 2018 ACPC and the strongest openly available agent in HUNL. OpenStack is a high-level poker AI integrated in OpenHoldem, a replica AI version of DeepStack. The experimental configurations are as follows.

For the first three rounds of the game, DecisionHoldem prioritizes using blueprint strategies when making decisions. For off-tree nodes, DecisionHoldem starts a real-time search. For the first two rounds of poker (preflop, flop), the real-time search iterations are 6,000 times; for the third round (turn), the real-time search iterations are 10,000 times. 

While for the last round (river), DecisionHoldem employs the safe depth-limited subgame solving algorithm for real-time search with 10,000 iterations directly.

\begin{figure*}[ht]
	\centering
	\includegraphics[width=1\linewidth]{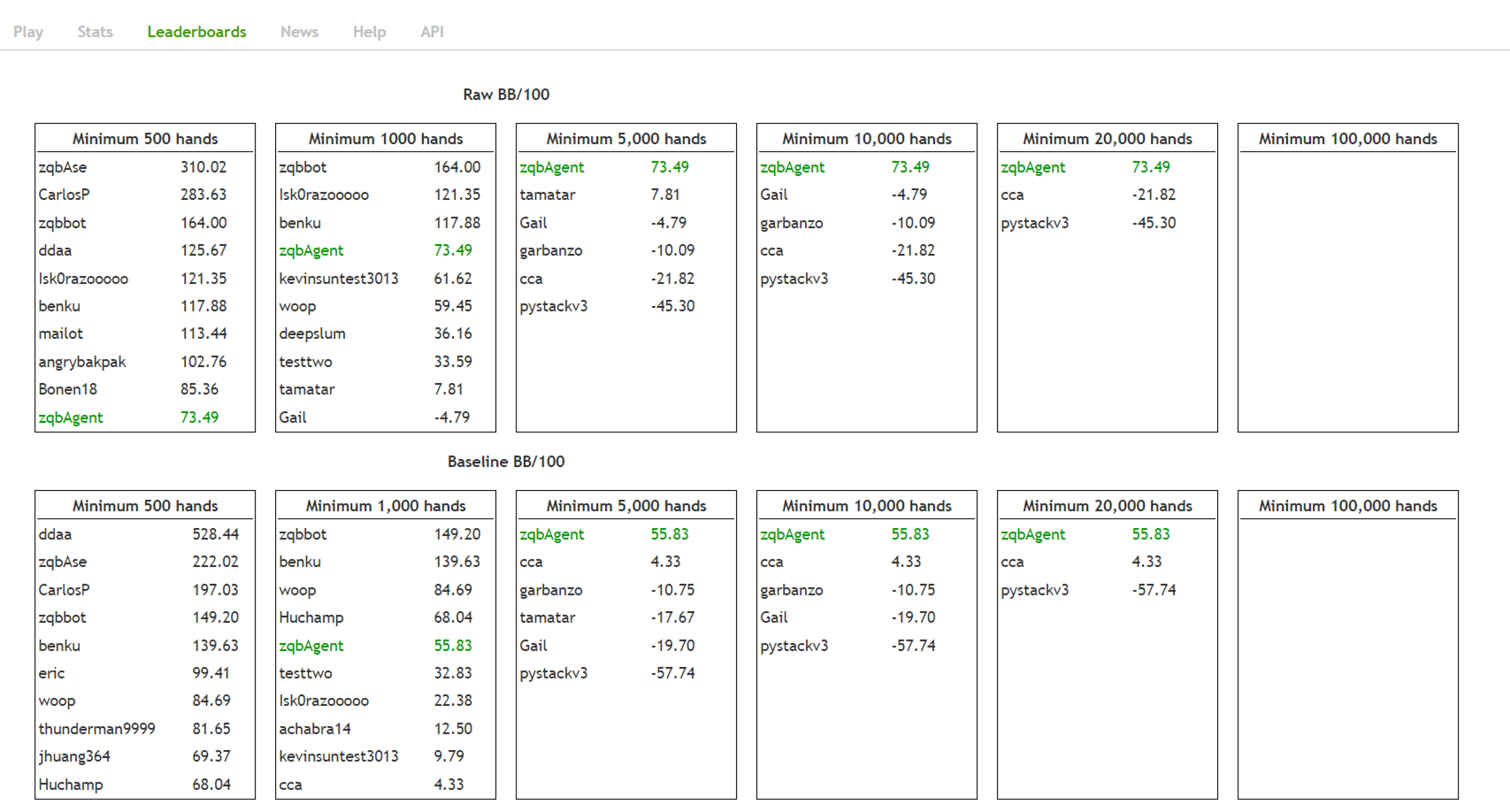}\caption{DecisionHoldem's ranking on the Slumbot leaderboard 
		.}
	\label{figure2}
\end{figure*}

\begin{figure}[ht]
	\centering
	\includegraphics[width=1\linewidth]{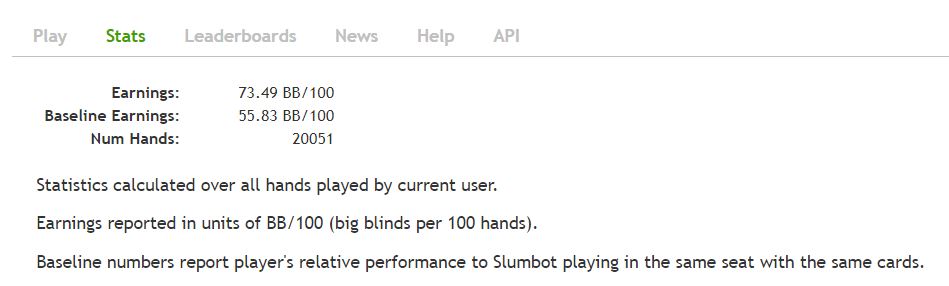}\caption{Statistics for DecisionHoldem vs. Slumbot.}
	\label{figure3}
\end{figure}

In approximately 20,000 games against Slumbot, DecisionHoldem's average profit is more remarkable than 730 mbb/h (one-thousandth big blind per round). It ranked first in statistics
 (DecisionHoldem's name on the leaderboard is zqbAgent\footnote{https://github.com/ericgjackson/slumbot2017/issues/11}), as the Figure \ref{figure2} and \ref{figure3}. With approximately 2,000 games against OpenStack, DecisionHoldem's average profit is greater than 700 mbb/h, and the competition records are available in the Github repository of DecisionHoldem.

\section{Conclusions}

This paper introduces the safe depth-limited subgame solving algorithm with the exploitability guarantee. It achieves the outstanding AI DecisionHoldem for HUNL with the proposed subgame solving algorithm for real-time search and suitable abstraction methods for blueprint strategy. DecisionHoldem defeats the current typical public high-level poker AI, namely Slumbot and OpenStack. To our best knowledge, DecisionHoldem is the very first open-source high-level AI for HUNL. Meanwhile, we also provide toolkits against Slumbot and OpenStack, and a platform to play DecisionHoldem with humans to assist researchers in conducting further research.

\bibliographystyle{named}
\bibliography{ijcai22}	

\end{document}